\pdfoutput=1
\documentclass[11pt]{article}

\usepackage[final]{acl}

\usepackage{times}
\usepackage{latexsym}

\usepackage[T1]{fontenc}
\usepackage[utf8]{inputenc}

\usepackage{adjustbox}

\usepackage{microtype}

\usepackage{inconsolata}

\usepackage{tikz}
\usetikzlibrary{bayesnet}
\usepackage{amsmath}
\usepackage{amssymb}
\usepackage{subcaption}
\usepackage{bm}

\usepackage{todonotes}
\definecolor{mintgreen}{RGB}{152, 255, 152}
\makeatletter
\newcommand*\iftodonotes{\if@todonotes@disabled\expandafter\@secondoftwo\else\expandafter\@firstoftwo\fi}  %
\makeatother

\usepackage{cleveref}
\crefname{section}{\S}{\S\S}
\crefname{table}{Tab.}{Tabs.}
\crefname{figure}{Fig.}{Figs.}
\crefname{subfigure}{Fig.}{Figs.}
\crefname{algorithm}{Alg.}{Algorithms}
\crefname{equation}{Eq.}{Eqs.}
\crefname{example}{Example}{Examples}
\crefname{fact}{Fact}{Facts}
\crefname{appendix}{App.}{Apps.}
\crefname{theorem}{Thm.}{Thms.}
\crefname{reTheorem}{Thm.}{Thms.}
\crefname{reCorollary}{Cor.}{Cors.}
\crefname{reTheoremSub}{Thm.}{Thms.}
\crefname{aquestion}{Question}{Questions}
\crefname{assumption}{Assumption}{Assumptions}
\crefname{lemma}{Lem.}{Lems.}
\crefname{claim}{Claim}{Claims}
\crefname{reLemma}{Lem.}{Lems.}
\crefname{proposition}{Prop.}{Props.}
\crefname{chapter}{Chapter}{Chapters}
\crefname{line}{line}{lines}
\crefname{principle}{Principle}{Principles}
\crefname{definition}{Def.}{Defs.}
\crefname{corollary}{Cor.}{Cors.}
\crefname{Exercise}{Exercise}{Exercises}
\crefformat{section}{\S#2#1#3}

\newcommand{\mcgill}{1}
\newcommand{\ethz}{2}
\newcommand{\cifar}{3}

\newcommand{\MI}{\text{MI}}
\newcommand{\PC}{\text{PC}}

\newcommand{\word}{w}
\newcommand{\lemma}{\ell}
\newcommand{\slot}{\sigma}
\newcommand{\slots}{\mathcal{S}}
\newcommand{\alphabet}{\Sigma}
\newcommand{\inflector}{\iota}
\newcommand{\lexicon}{\mathcal{L}}
\newcommand{\lexeme}{\bm{\ell}}

\title{Correlation Does Not Imply Compensation:\\Complexity and Irregularity in the Lexicon}
\author{
   Amanda Doucette$^{\mcgill}$\quad Ryan Cotterell$^{\ethz}$\quad Morgan Sonderegger$^{\mcgill}$ \quad Timothy J. O'Donnell$^{\mcgill, \cifar}$\\
   $^{\mcgill}$Dept. of Linguistics, McGill University \quad $^{\ethz}$Dept. of Computer Science, ETH Zurich \\ 
   $^{\cifar}$Canada CIFAR AI Chair, Mila \\
    \texttt{\href{mailto:amanda.doucette@mail.mcgill.ca}{amanda.doucette@mail.mcgill.ca}} \quad \texttt{\href{mailto:ryan.cotterell@inf.ethz.ch}{ryan.cotterell@inf.ethz.ch}} \\ \texttt{\href{mailto:morgan.sonderegger@mcgill.ca}{morgan.sonderegger@mcgill.ca}} \quad \texttt{\href{mailto:timothy.odonnell@mcgill.ca}{timothy.odonnell@mcgill.ca}}
}

\begin{document}
\maketitle
\begin{abstract}

    It has been claimed that within a language, morphologically irregular words are more likely to be phonotactically simple and morphologically regular words are more likely to be phonotactically complex.
    This inverse correlation has been demonstrated in English for a small sample of words, but has yet to be shown for a larger sample of languages.
    Furthermore, frequency and word length are known to influence both phonotactic complexity and morphological irregularity, and they may be confounding factors in this relationship.
    Therefore, we examine the relationships between all pairs of these four variables both to assess the robustness of previous findings using improved methodology and as a step towards understanding the underlying causal relationship.
    Using information-theoretic measures of phonotactic complexity and morphological irregularity \citep{pimentel-etal-2020-phonotactic, wu-etal-2019-morphological} on 25 languages from UniMorph, we find that there is evidence of a \emph{positive} relationship between morphological irregularity and phonotactic complexity within languages on average, although the direction varies within individual languages.
    We also find weak evidence of a negative relationship between word length and morphological irregularity that had not been previously identified, and that some existing findings about the relationships between these four variables are not as robust as previously thought.\footnote{To appear in \textit{Proceedings of the Society for Computation in Linguistics 2024}. Code is available at \url{https://osf.io/ax78p/}.}
\end{abstract}

\section{Introduction}

The \textit{compensation hypothesis} \citep{martinet1955economie, hockett1955manual} states that as a language increases in complexity in one area, another must decrease in complexity to compensate.
A compensatory relationship could exist either \textit{within} a language (i.e., words that are more complex in one way are less complex in another), or \textit{across} languages (i.e., an entire lexicon that is more complex in one way is less complex in another). One such compensatory relationship has been proposed between morphological irregularity and phonotactic complexity. \citet{hay2000causes}, \citet{hay2003phonotactics}, and \citet{burzio2002missing} argue that words within a language with irregular morphology tend to be phonotactically simple, while words with regular morphology tend to be phonotactically complex.

Although there is some evidence for this relationship in English (reviewed below), the existence of a correlation does not imply \emph{compensation}, which we take to mean that an increase in one variable directly causes a decrease in the other \citep[][\S{1.5}]{pearl2016causal}. While we may observe a correlation between morphological irregularity and phonotactic complexity, it is possible that there is in fact no direct causal relationship between them. For example, they could share a common cause such as word frequency \citep[][\S{2.2}]{pearl2016causal}. The effect could also be mediated through a third mediator variable \citep[][\S{3.7}]{pearl2016causal}, as has been argued for the relationship between phonotactic complexity and frequency \citep{mahowald}.

Therefore, to assess the relationship between morphological irregularity and phonotactic complexity, we need to examine any other variables they may be related to. Previous work suggests that both morphological irregularity and phonotactic complexity are correlated with word frequency. \citet{wu-etal-2019-morphological} showed that morphological irregularity positively correlates with frequency, and \citet{mahowald} showed that phonotactic complexity inversely correlates with frequency after controlling for word length. Phonotactic complexity is also known to be correlated with word length, with longer words conveying less information per phoneme \citep{pimentel-etal-2020-phonotactic}, and %Furthermore, 
more frequent words tend to be shorter \citep{zipf, piantadosi2011word, piantadosi, pimentel-etal-2023-revisiting}.\looseness=-1

While there is evidence supporting a relationship between some pairs of these four variables---morphological irregularity, phonotactic complexity, word length, and frequency---there is reason to be uncertain about the existence and direction of a correlation between others, whether the correlation holds within or across languages, and what other variables need to be controlled for to accurately assess the effect. The pairwise relationships between these variables have not yet been examined on a single data set of many languages, and some relationships have only been examined using orthographic rather than phonetic transcriptions. Therefore, in addition to examining the relationship between phonotactic complexity and morphological irregularity, we will also examine the relationships between all other pairs of variables in this set.

We find that within languages, there is a positive effect of phonotactic complexity on morphological irregularity after controlling for word length and frequency. Across languages, we find no consistent effect.
We replicate previous findings of a negative effect of word length on frequency, and of a positive effect of frequency on morphological irregularity. We also find a negative effect of frequency on phonotactic complexity, although not as robust as previously suggested. Our results for the relationship between phonotactic complexity and word length complicate previous results: We find a positive effect for one data set and a negative effect for another. Finally, we present a novel analysis of the effect of word length on morphological irregularity, and find a negative effect in most languages.\looseness=-1

\section{Background}

For each pair of variables, we summarize previous work demonstrating a correlation, any theoretical arguments supporting a positive, negative, or no relationship between the two, and what other variables must be controlled for to examine a potential causal relationship.

\subsection{Phonotactic Complexity vs. Morphological Irregularity}

It has been hypothesized that within a language, phonotactic complexity is negatively  correlated  with morphological irregularity. \citet{hay2003phonotactics}, \citet{hay2000causes}, and \citet{burzio2002missing} argued that for English, words that are phonotactically complex are more likely to be morphologically regular. For example, \textit{dreamed} is morphologically regular, but contains the unusual consonant cluster [md], while  \textit{went} is morphologically irregular, but has regular phonotactics.
While this relationship has not been  examined in other languages, there are several reasons to suspect a negative correlation as a universal tendency.
First, low-probability phonotactic junctures can facilitate morphological decomposition, as argued by \citet{hay2000causes}, who found that for a set of 12 English affixes the proportion of words creating an illegal phonotactic juncture was predictive of morphological productivity. Second, as argued by \citet{burzio2002missing}, irregular forms are more likely to be memorized, while regular forms are constructed from individual morphemes. If phonotactically simple words are easier to store in memory, phonotactic complexity should be inversely correlated with morphological irregularity.

There are also reasons to suspect no relationship between  morphological irregularity and phonotactic complexity, related to the limitation that all previous work considers only a  small set of words.
Morphological and phonotactic processes could apply independently,
and previously observed significant correlations could be statistical accidents due to small sample size.
Indeed, responding to \citet{hay2000causes}, \citet{plag2002role} found no correlation between morphological irregularity and phonotactic complexity in a different sample of 12 English affixes.
Alternatively, morphological irregularity and phonotactic complexity could be independent conditional on a third common cause or mediator variable, with which they are both correlated.  This could result in a statistically significant correlation, while there is no causal relationship in reality.

One such common cause that could result in a positive observed correlation between phonotactic complexity and morphological regularity is word age. \citet{hay2003phonotactics} note that highly productive affixes are regularly used in creating new words. Thus, new words in a language will tend to have regular morphology, and it seems plausible they will also tend to have regular phonotactics. As the language changes over time, what is considered regular will also change, resulting in a positive correlation: Older words will have irregular morphology and high phonotactic complexity, while newer words will have regular morphology and low phonotactic complexity.
Other possible common causes include word length and frequency: Previous work  demonstrates correlations between frequency and phonotactic complexity, frequency and morphological irregularity, and word length and phonotactic complexity. A negative effect of word length on morphological irregularity is also  plausible, and will be established in our data.   Both word frequency and word length are therefore common causes  that should be controlled for in assessing the relationship between  phonotactic complexity and morphological irregularity.

\subsection{Phonotactic Complexity vs. Length}

\citet{pimentel-etal-2020-phonotactic} demonstrated a strong negative correlation between phonotactic complexity and average word length both across and within 106 languages. \citet{pellegrino2011cross} suggest that this compensation is the result of a linguistic universal:
The rate of information in every language is very similar, with the amount of information per word roughly constant. Thus, longer words should have less information per phoneme \citep{coupe2019, meister-etal-2021-revisiting}.
While previous work (discussed below) suggests that frequency is a common cause of both phonotactic complexity and word length, and should be controlled for in this analysis, we will not control for it in line with our goal of replicating previous studies using a single dataset.\footnote{However, preliminary models show that controlling for frequency only has a minimal impact on results.}\looseness=-1

\subsection{Morphological Irregularity vs. Frequency}

A positive correlation between morphological irregularity and word frequency has been observed in English \citep{marcus1992overregularization, bybee1985morphology}, but this correlation was questioned by
\citet{fratini2014frequency} and \citet{yang2016price}.
In a larger set of 21 languages, \citet{wu-etal-2019-morphological} found a positive correlation between morphological irregularity and frequency. These correlations were found to be more robust when irregularity was considered as a property of lemmas rather than individual words.
A potential mechanism is described by \citet{hay2003phonotactics}:
More frequent words are more likely to be accessed as whole words, and less frequent words are more likely to be parsed into their component morphemes. Because irregulars are more likely to be accessed as whole words in memory, there will be a positive correlation between frequency and morphological irregularity.
In contrast, if we assume that lexicons are optimized for efficient communication, i.e., more frequent words should be less morphologically complex \citep{zipf}, we would expect frequent words to have regular morphology, i.e., a negative correlation.\looseness=-1

\subsection{Phonotactic Complexity vs. Frequency}

A consequence of \citeposs{zipf} hypothesis that the most frequent words in a language should require the least effort is that even within words of the same length, the most frequent ones should be easiest to produce and understand. This suggests that more frequent words should have lower phonotactic complexity. After controlling for word length, this is exactly what \citet{mahowald} found in a study of 96 languages, using orthographic probabilities from Wikipedia as a proxy for phonotactic complexity. However, orthography can differ significantly from pronunciation---this correlation has not been confirmed with phonotactic probabilities from phonetic transcriptions. Following \citet{mahowald}, we will control for word length as a potential mediator in the relationship between phonotactic complexity and word frequency.

\subsection{Morphological Irregularity vs.\ Length}

We are not aware of previous  work on the relationship between morphological irregularity and word length, although it is intuitively plausible that one influences the other.  For example, a negative correlation within a language could arise because regular inflectional morphology involves combining multiple morphemes, causing words with regular morphology to be longer. Previous work also implies that frequency has an effect on both morphological irregularity and word length. Therefore, we control for frequency as a common cause in assessing the potential relationship between morphological irregularity and word length.

\subsection{Length vs. Frequency}

\citet{zipf} observed that the most frequent words in a language tend to be short. Since then, the inverse relationship between word length and frequency has been studied in depth and found to follow
Zipf's law extremely systematically (see \citealp{piantadosi} for a review), although it is unclear whether word length correlates more strongly with surprisal \citep{piantadosi2011word} or frequency \citep{meylan2021}.

\section{Methods}
\subsection{Data}

Our morphological data comes from the UniMorph project, a database of morphologically annotated corpora for 182 languages \citep{batsuren-etal-2022-unimorph}. Each inflected form is annotated with its lemma (the lexical meaning) and a set of morphological features; \textit{walked} would be annotated as [\textsc{verb}; \textsc{singular}; \textsc{past}], for example.
While UniMorph provides data for a large set of languages in a universal schema, it does not provide phonetic transcriptions.
Because we are interested in how morphology interacts with phonotactics, we use grapheme to phoneme models from Epitran \citep{Mortensen-et-al:2018} to convert orthographic transcriptions to IPA. Languages with no available Epitran model were excluded from our analyses.

For training models of phonotactic complexity, we use NorthEuraLex \citep{northeuralex}, a database of phonetic transcriptions of 1,016 basic concepts for 107 Northern Eurasian languages also studied by \citet{pimentel-etal-2020-phonotactic}.
For languages that are not included in NorthEuraLex, we use WikiPron \citep{lee-etal-2020-massively}, a database of pronunciation dictionaries from Wiktionary.
Languages not in either NorthEuraLex or WikiPron are excluded.

Frequency data is retrieved from Wikipedia,\footnote{Retrieved 4/16/23 from https://dumps.wikimedia.org.} and calculated as log count per million words.
Following \citet{wu-etal-2019-morphological}, we exclude all forms with zero frequency, which can differ by orders of magnitude in their true frequency \citep{baayen2001word}, and we exclude 15 languages where the average probability of the morphological irregularity model predicting the correct surface form is below $0.75$. The UniMorph dataset for each language varies in size from 77 to 50,284,287 forms and 37 to 824,074 lemmas. Many of the excluded languages are those with smaller datasets, where the model does not have enough information to accurately predict surface forms.\looseness=-1

The languages included in our analysis are: Albanian, Amharic, Azerbaijani, Catalan, Chewa, Czech, Dutch, English, French, German, Hungarian, Italian, Kazakh, Khalka Mongolian, Polish, Portuguese, Romanian, Russian, Serbo-Croatian, Spanish, Swedish, Turkish, Ukrainian, Uzbek, and Zulu.
Further details about datasets used can be found in \cref{sec:appendix_a}.\looseness=-1

\subsection{Morphological Irregularity Models}

While a binary distinction between regular and irregular morphology is useful in many theories of grammar, a more fine-grained quantitative measure of irregularity is needed to examine the potential relationships with other variables we are interested in.
For example, an English speaker might judge a verb like \textit{walked} to be more regular than \textit{sang}, which is, in turn, more regular than \textit{went}.
\citet{wu-etal-2019-morphological} define an information theoretic measure of irregularity that captures such intuitions, and is applicable across languages.

This measure, called \textit{degree of morphological irregularity}, abbreviated as $\MI$, is defined using a probabilistic model.
Let $\alphabet$ be an alphabet of symbols\footnote{The symbols could be graphemes or phonemes, depending on the nature of the data annotation.} and let $\slots$ be a finite set of morphological features, e.g., those provided by the UniMorph dataset. We define a lemma $\lemma$ as the citation form of a word $\word \in \alphabet^*$, and a partial function $\inflector \colon \alphabet^* \times \mathcal{P}(\slots)  \rightarrow \alphabet^*$ that, when defined, maps a pair of a lemma and a set of morphological features to an inflected surface form, i.e., $(\lemma, \slot) \mapsto \word$. The inflector $\inflector$ is assumed to only operate on \emph{known} lemma and feature combinations. Furthermore, we define a lexeme $\lexeme$ as the set of inflected surface forms associated with a lemma, $\{\inflector(\lemma, \slot) \mid \slot \in \mathcal{P}(\slots)\}$.
Thus, to get at a notion of morphological irregularity, we also require a probabilistic inflection model $p(\word \mid \lemma, \slot, \lexicon_{-\lexeme})$, a probability distribution over $\alphabet^*$ conditioned on a lemma $\lemma \in \alphabet^*$, a slot $\slot \in \mathcal{P}(\slots)$, and the set of forms in the lexicon of the target language with the target lexeme removed $\lexicon_{-\lexeme}$, that tells us which forms in $\alphabet^*$ are probable inflected surface forms for the lemma $\lemma$ with morphological features $\slot$.
The distribution $p(\word \mid \lemma, \slot, \lexicon_{-\lexeme})$  essentially corresponds to a wug-test probability \citep{berko1958child}, i.e., it tells us the likelihood of the model predicting the correct inflected form of a word it has never seen.
To make the probabilities given by $p(\word \mid \lemma, \slot, \lexicon_{-\lexeme})$ more interpretable, \citet{wu-etal-2019-morphological} use the negative log odds of probability of the correct surface form, i.e.,
\begin{equation}\label{eq:mi}
    \MI(\word, \lemma, \slot) = -\log  \frac{p(\word \mid \lemma, \slot, \lexicon_{-\lexeme})}{1 - p(\word \mid \lemma, \slot, \lexicon_{-\lexeme})}
\end{equation}
We can interpret $\MI(\word, \lemma, \slot)$ as follows.
We achieve a $\MI(\word, \lemma, \slot)$ of 0 if the probability of the correct surface form is exactly 0.5, a negative $\MI(\word, \lemma, \slot)$ when a surface form is more predictable, and a positive $\MI(\word, \lemma, \slot)$ when the form is less predictable.

Morphological irregularity can be considered either a property of an individual word, as in \Cref{eq:mi}, or as a property of an entire lemma.
We calculate the $\MI$ of a lemma as the mean $\MI$ score of all surface forms associated with the lemma, i.e.,
\begin{equation}\label{eq:mi_lem}
    \MI(\lemma) = \frac{1}{|\slots|}\sum_{\slot \in \slots} \MI(\inflector(\lemma, \slot), \lemma, \slot)
\end{equation}

\paragraph{Estimating $p(\word \mid \lemma, \slot, \lexicon_{-\lexeme})$ from data.}
Because the probability distribution $p(\word \mid \lemma, \slot, \lexicon_{-\lexeme})$ is conditioned on a lexicon with the target lexeme removed, the most accurate estimate of $\MI$ would require training a separate model for each target lexeme in each language.
In practice, $\MI$ is estimated by training models on a language with a set of lexemes removed, rather than just one. We train neural network models on the UniMorph data using code from \citet{wu-etal-2019-morphological}, which implements a monotonic hard attention string-to-string induction model described by \citet{wu-cotterell-2019-exact}. In our experiments, we use the same model architecture and training parameters as \citet{wu-etal-2019-morphological}, and split the lemmas for each language into thirty sets.\looseness=-1

\subsection{Phonotactic Complexity Models}
Similar to to our estimation of morphological irregularity, to estimate phonotactic complexity, we take a probabilistic approach.
Following \citet{pimentel-etal-2020-phonotactic}, we consider a probability distribution $p(\word \mid \lexicon_{-\word})$ over $\alphabet^*$, conditioned on $\lexicon_{-\word}$, the set of forms in the lexicon of the target language with the target word removed.
Then, given a word $\word \in \alphabet^*$, we define the degree of phonotactic complexity, abbreviated as $\PC$, as follows
\begin{equation}
    \PC(\word) = - \frac{\log p(\word \mid \lexicon_{-\word})}{|\word|}
\end{equation}
where $|\word|$ is the length of the word $\word$.
Like the degree of morphological irregularity, $\PC$ is a surprisal-based metric that lends itself to easy interpretation.
Specifically, if $\PC(\word)$ is lower, it means that $\word$ is less surprising and therefore more regular.

\paragraph{Estimating $p(\word \mid \lexicon_{-\word})$ from data.}
The distribution $p(\word \mid \lexicon_{-\word})$ is a hypothetical construct that tells us the probability of an \emph{unknown} word.
When we estimate $p(\word \mid \lexicon_{-\word})$ from data, we cannot use the estimated distribution to judge the complexity of those words in the training data. We split training data for each language into ten sets and train ten models, each with one set held out. $\PC$ is evaluated on the held-out set for each model. We use the model architecture and training procedure used in code provided by \citet{pimentel-etal-2020-phonotactic}, which implements a character-level LSTM  \citep{hochreiter1997long} language model with each phoneme represented by a set of phonetic features from Phoible \citep{moran2014phoible}.\looseness=-1

\begin{table}
    \begin{adjustbox}{width=\columnwidth}
        \begin{tabular}{llll}
            \hline
            $\mathbf{Y}$ & $\mathbf{X}$ & \textbf{Controls}           & \textbf{Rand. Effs.} \\
            \hline
            MI           & PC           & \underline{FR} + mean(PC) + & PC + FR + WL         \\
                         &              & \underline{WL} + mean(WL)   &                      \\
            PC           & WL           & mean(WL)                    & WL                   \\
            MI           & FR           & --                          & FR                   \\
            PC           & FR           & \underline{WL} + mean(WL)   & FR + WL              \\
            MI           & WL           & \underline{FR} + mean(WL)   & WL + FR              \\
            WL           & FR           & --                          & FR                   \\
            \hline
        \end{tabular}
    \end{adjustbox}
    \caption{Controls and random effects included in regression models of properties $X$ and $Y$ of words  (one row per word): MI = morphological irregularity; PC = phonotactic complexity; WL = word length; FR = frequency. Mean($X$) is a language's average value of $X$, across all its words.   In lme4 syntax, models using words from all languages are: \texttt{Y $\sim$ X + Controls + (1 + Random Effects | language)}, and individual language models are: \texttt{Y $\sim$ X + UnderlinedControls}.}
    \label{tab:regression}
\end{table}

\subsection{Analysis}

Following previous studies on the interactions between phonotactic complexity, morphological irregularity, word length, and frequency, we report regression coefficients for each language and each pair of variables, controlling for any necessary variables as described in \cref{tab:regression}---these are analogous to partial correlations between the variables of interest.
The $p$-values have been adjusted for multiple comparisons using the Benjamini-Hochberg method.
We also perform a linear mixed effects regression analysis across languages for each pair of variables, including random intercepts and slopes for the effect of language. The means of the dependent variable and controls within each language are also included as an additional predictor, to separate across-language effects from within-language effects (see \citealp[][\S{8.10.2.3}]{sonderegger2023regression} and \citealp{antonakis2021ignoring}). Because languages should not differ in mean word frequency, this predictor is excluded. All predictors were standardized, and log counts-per-million were used for frequencies.

\begin{figure*}
    \begin{subfigure}[t]{0.45\textwidth}
        \centering
        \includegraphics[width=\textwidth]{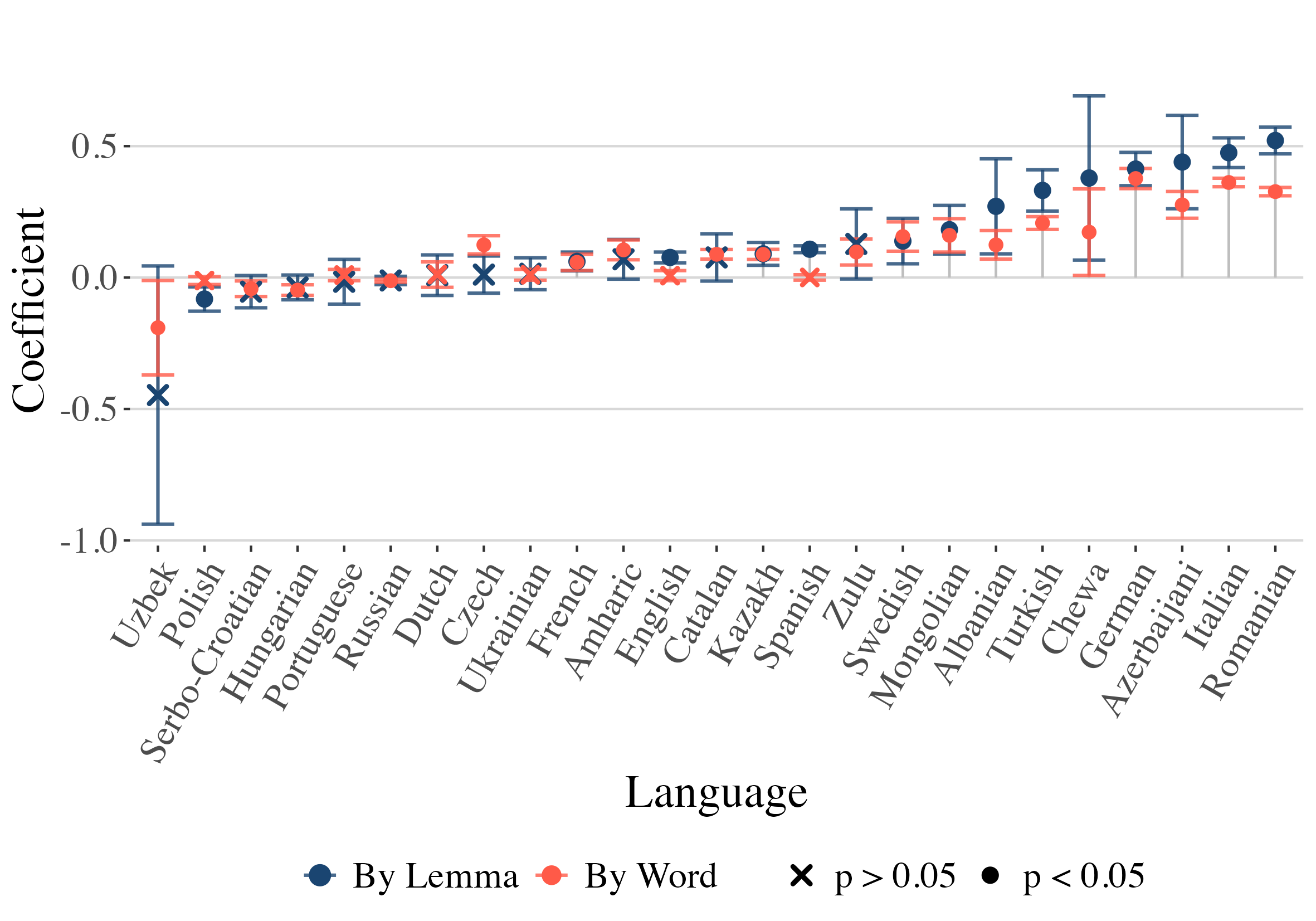}
        \caption{Regression coefficients by language, grouped by lemma and by word, with 95\% CIs.}
        \label{fig:mi_pc_corr}
        \vspace{5pt}
    \end{subfigure}
    \hfill
    \begin{subfigure}[t]{0.45\textwidth}
        \centering
        \includegraphics[width=\textwidth]{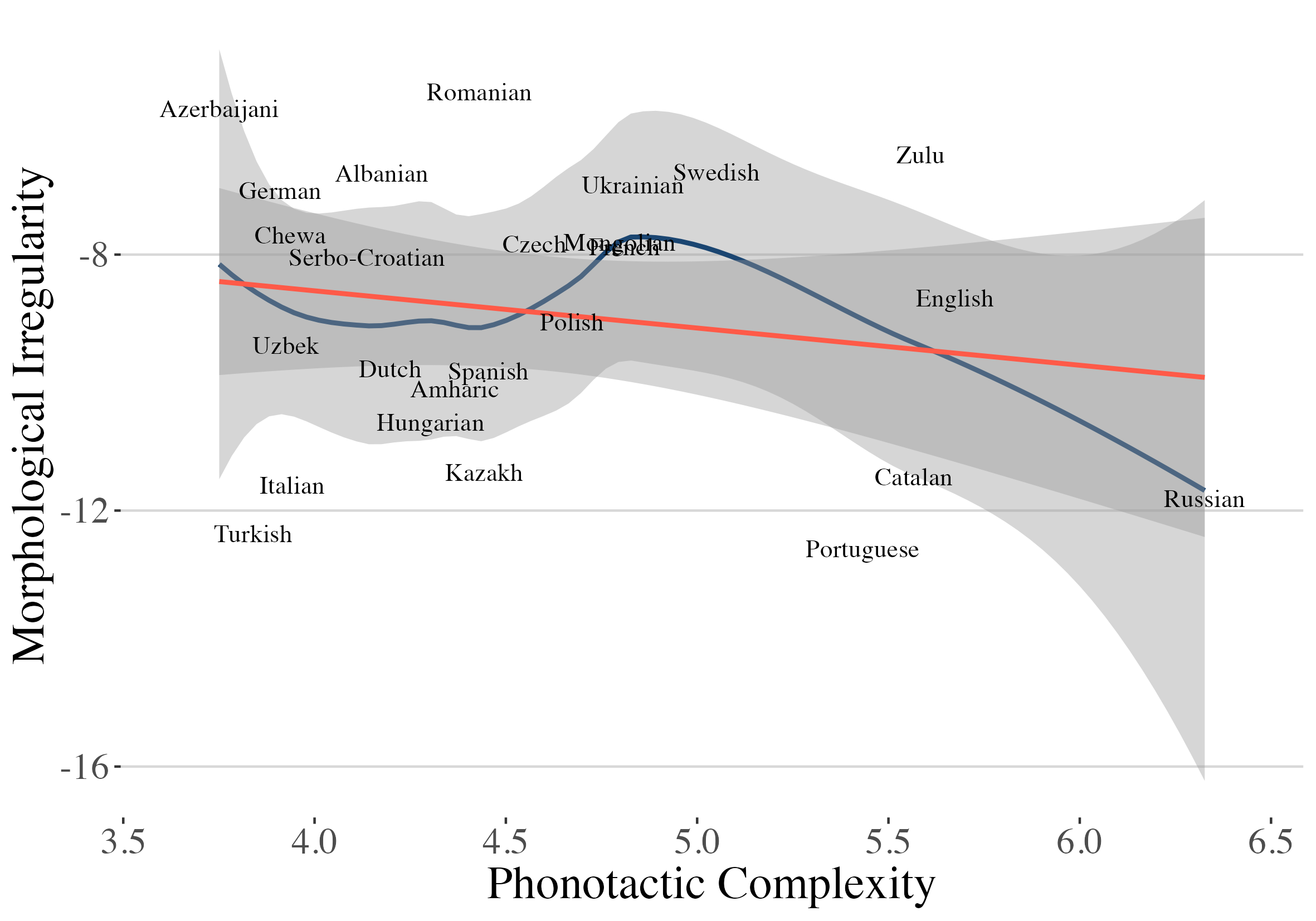}
        \caption{By-language means, with linear (red/light) and LOESS (blue/dark) smoothers.}
        \label{fig:mi_pc_langs}
    \end{subfigure}
    \caption{Phonotactic complexity and morphological irregularity.}
\end{figure*}

\begin{figure*}[ht]
    \begin{subfigure}[t]{0.45\textwidth}
        \centering
        \includegraphics[width=\textwidth]{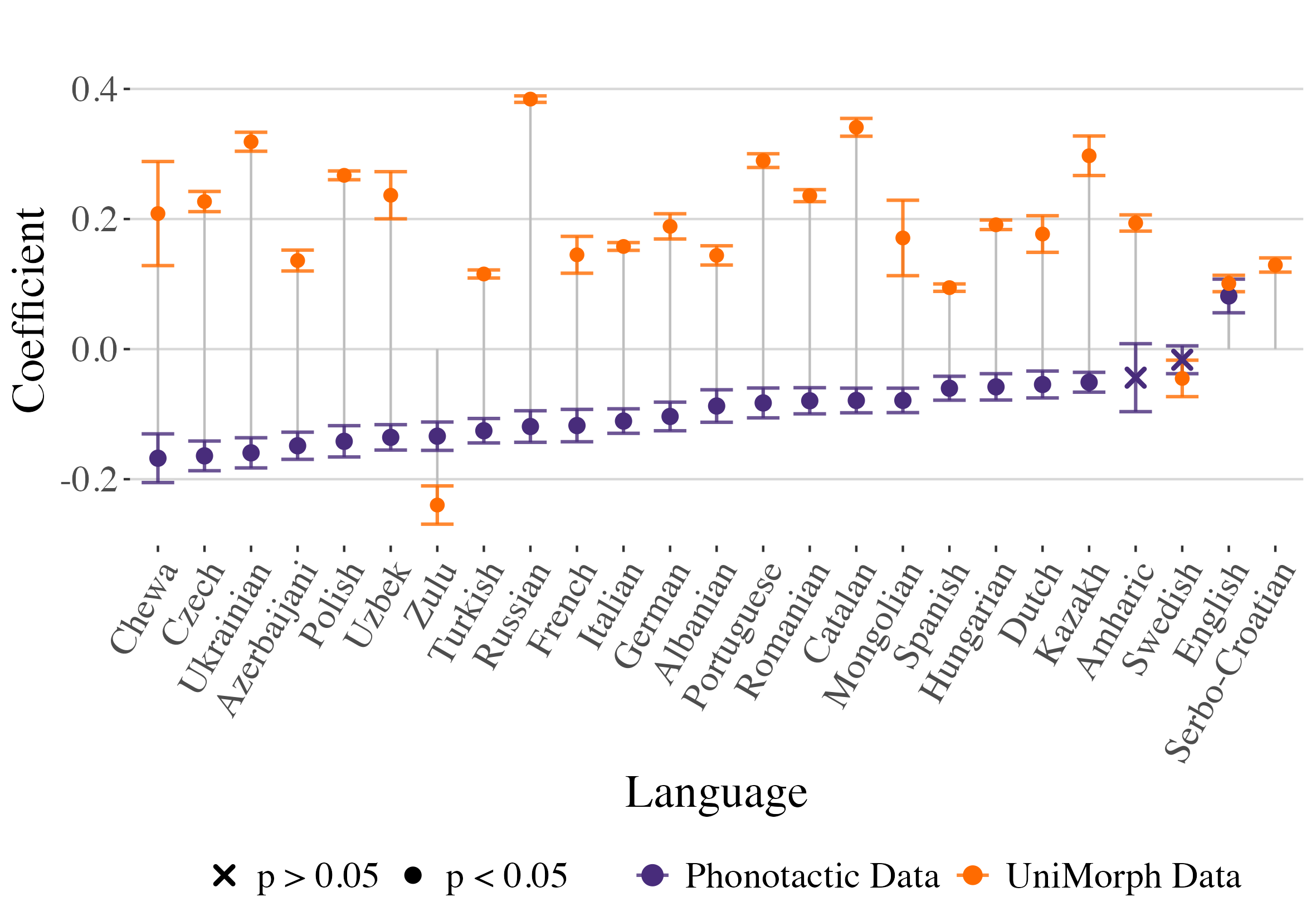}
        \caption{Regression coefficients by language, for phonotactic and UniMorph data, with 95\% CIs.}
        \label{fig:pc_wl_corr}
        \vspace{15pt}
    \end{subfigure}
    \hfill
    \begin{subfigure}[t]{0.45\textwidth}
        \centering
        \includegraphics[width=\textwidth]{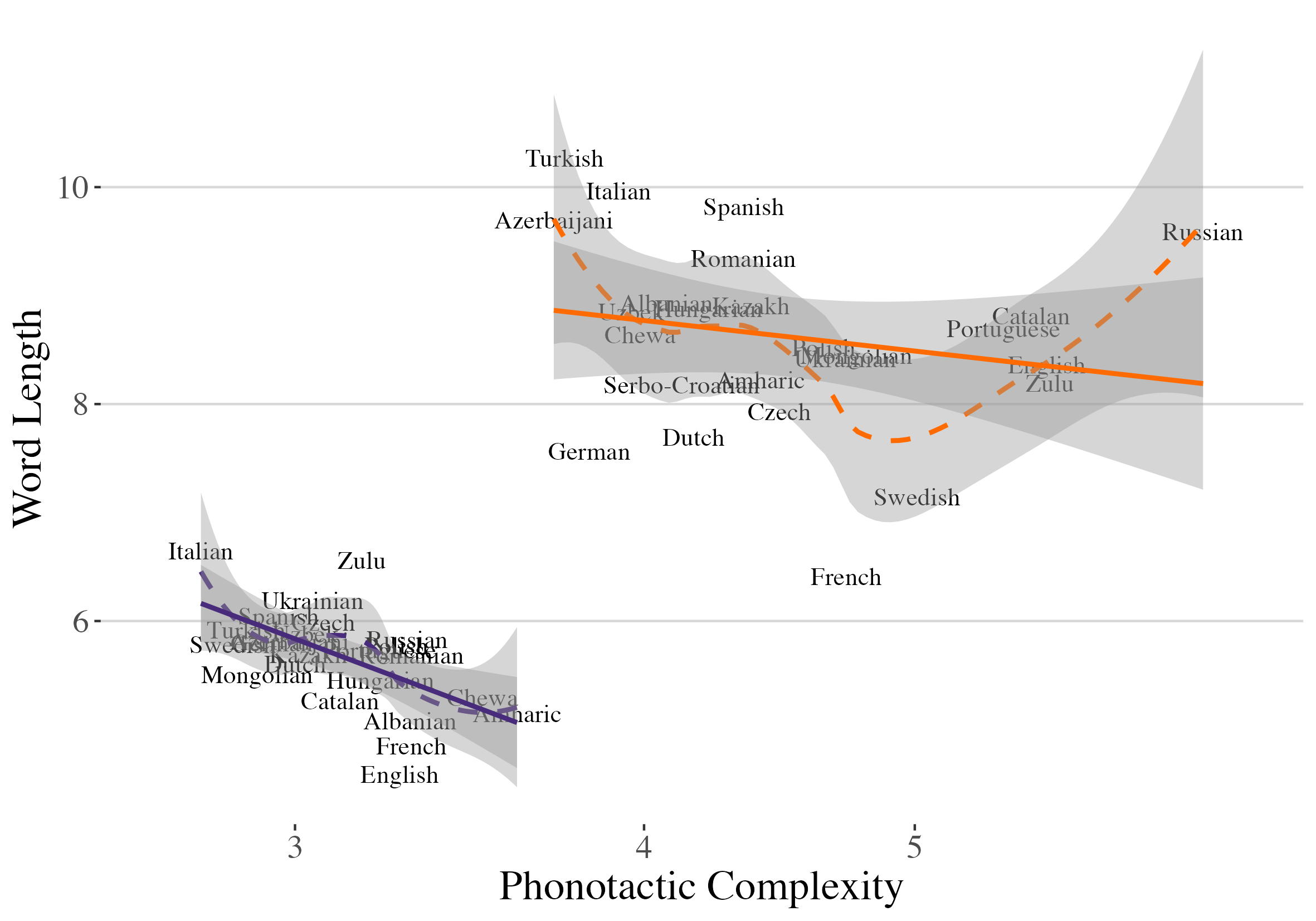}
        \caption{By-language means for phonotactic data (purple/dark) and UniMorph data (orange/light), with linear (solid) and LOESS (dotted) smoothers.}
        \label{fig:pimrep-1}
    \end{subfigure}
    \caption{Phonotactic complexity and word length.}
\end{figure*}

For morphological irregularity analyses (where $Y$ = MI in \cref{tab:regression}), we also report a regression model with data grouped by lemma, where phonotactic complexity and word length are taken as the average within each lemma, frequency is the sum of frequencies within each lemma, and morphological irregularity is calculated according to \cref{eq:mi_lem}. Results for the effects of interest in these regression analyses are reported below. Plots of model predictions and raw data for select languages are shown in \cref{sec:appendix_b}.

\section{Results}

\subsection{Phonotactic Complexity and Morphological Irregularity}

Within languages, we find a positive effect of phonotactic complexity on morphological irregularity after controlling for word length and frequency of 0.10 (95\% CI [0.05, 0.15], $p < 0.001$, $\sigma = 0.188$)\footnote{We use $\sigma$ to refer to random effect standard deviation.\looseness=-1} and a non-significant ($p > 0.05$) effect of mean phonotactic complexity of -0.19 (95\% CI [-0.50, 0.12], $p = 0.224$). When grouped by lemma, there is an estimated effect of 0.14 (95\% CI [0.07, 0.21], $p < 0.001$, $\sigma = 0.261$) and a non-significant effect of mean phonotactic complexity of 0.00 (95\% CI [-0.36, 0.37], $p = 0.993$). Although a majority of languages have a positive effect, as shown in \cref{fig:mi_pc_corr}, some are negative or non-significant. Across languages, there is no evidence of a relationship between mean morphological irregularity and mean phonotactic complexity---we find a non-significant Spearman's correlation of $-0.045$ ($p = 0.832$), shown in \cref{fig:mi_pc_langs}.

\subsection{Phonotactic Complexity and Word Length}

As shown in \cref{fig:pc_wl_corr}, we find a positive relationship between phonotactic complexity and word length within the majority of languages in the UniMorph data set. We find a similar prediction from the linear model: the estimated effect is 0.18 (95\% CI [0.13, 0.23], $p < 0.001$, $\sigma = 0.123$). Across languages, we find no evidence of a correlation between mean phonotactic complexity and mean word length ($\rho = -0.333$, $p = 0.104$), as shown in \cref{fig:pimrep-1}. Similarly, the linear model estimates no significant effect of mean word length ($\beta = -0.40$, 95\% CI [$-0.95$, 0.14], $p = 0.145$).
The positive effects found in most languages are opposite to the direction predicted by \citet{pimentel-etal-2020-phonotactic} and \citet{pellegrino2011cross}. However, it is important to note that nearly all words in UniMorph are morphologically complex, while NorthEuraLex (the dataset used by \citet{pimentel-etal-2020-phonotactic}) contains mostly morphologically simple words. As previously noted, morpheme boundaries can create low-probability phonotactic junctures, resulting in higher phonotactic complexity. This suggests that the relationship between word length and phonotactic complexity may be dependent on morphological complexity. We also note that the UniMorph data set contains significantly longer words than NorthEuraLex, suggesting a potential non-linear effect.

To test this, we evaluate the phonotactic complexity of the NorthEuraLex and WikiPron data used to train the phonotactic models and fit models for each language. These results are shown in \cref{fig:pc_wl_corr}, where we can see that most languages show a negative effect in morphologically simple data, replicating the results of \citet{pimentel-etal-2020-phonotactic}. We find a Spearman's correlation coefficient of $-0.578$ ($p = 0.003$) between average $\PC$ and average word length across languages, as shown in \cref{fig:pimrep-1}.
These results suggest that the finding that phonotactic complexity is negatively correlated with word length only holds for either morphologically simple or relatively short words.

\subsection{Morphological Irregularity and Frequency}

\begin{figure}
    \centering
    \includegraphics[width=0.45\textwidth]{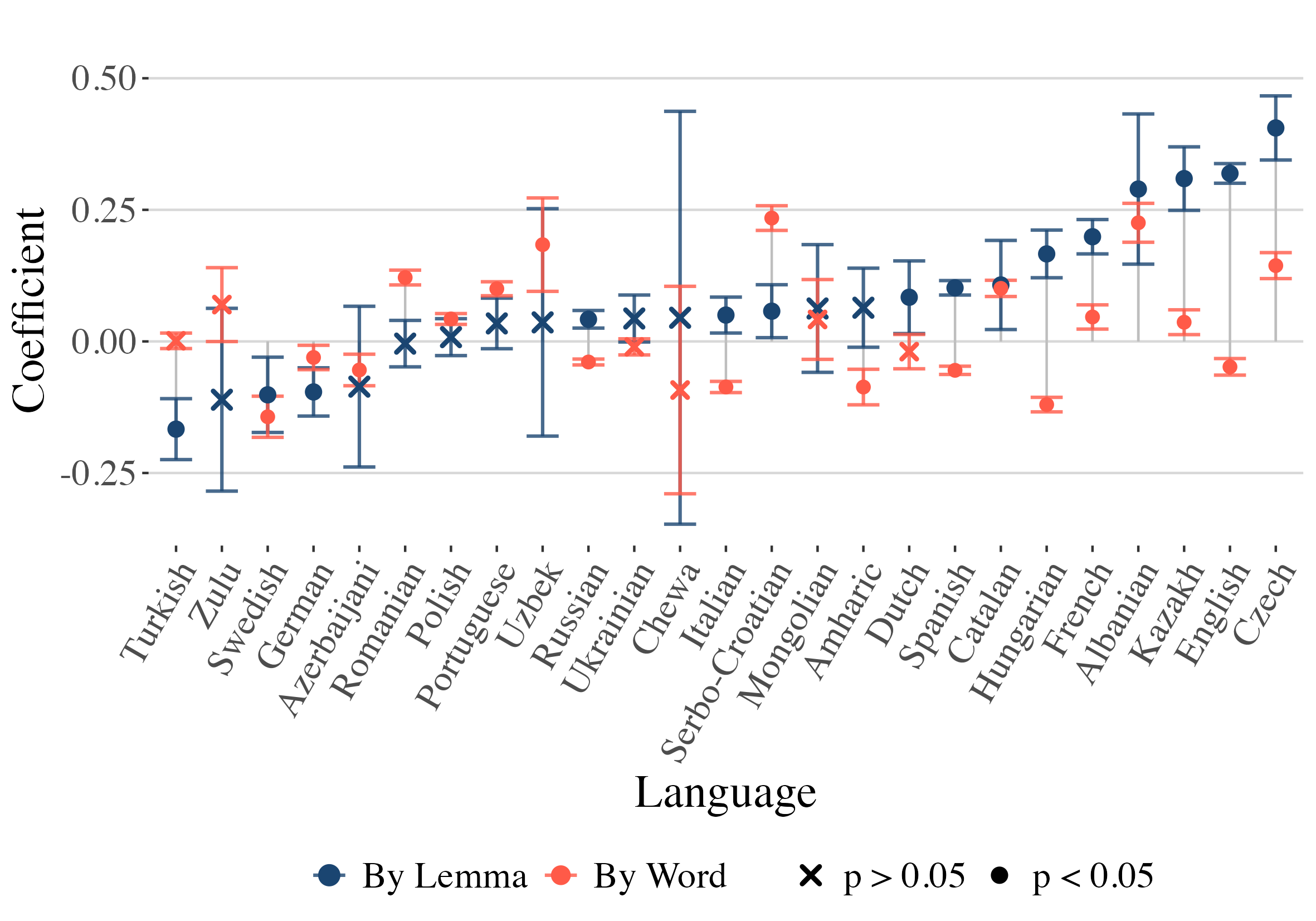}
    \caption{Morphological irregularity and frequency coefficients by language, grouped by lemma and by word, with 95\% CIs.}
    \label{fig:mi_fr_corr}
\end{figure}

On the one hand, when morphological irregularity is considered a property of individual words, we find that 10/25 languages have a significant positive effect of frequency on morphological irregularity, and 9/25 have a significant negative effect (\cref{fig:mi_fr_corr}). In the linear mixed-effects model, we find no significant effect ($\beta = 0.02$, 95\% CI [$-0.02$, 0.06], $p = 0.224$, $\sigma = 0.100$).
On the other hand, when morphological irregularity is considered a property of lemmas rather than individual words, we find that 12/25 languages have a positive effect and only 3/25 have a negative effect. These correlations are shown in \cref{fig:mi_fr_corr}. The linear mixed-effects model predicts a positive effect of 0.08 (95\% CI [0.02, 0.13], $p = 0.005$, $\sigma = 0.135$), consistent with those of \citet{wu-etal-2019-morphological} who examined the same relationship, although using the original orthographic transcriptions from UniMorph, rather than G2P transcriptions. Although the direction of correlation varies across individual languages, there is a tendency towards a positive correlation.

\subsection{Phonotactic Complexity and Frequency}

\begin{figure}
    \centering
    \includegraphics[width=0.45\textwidth]{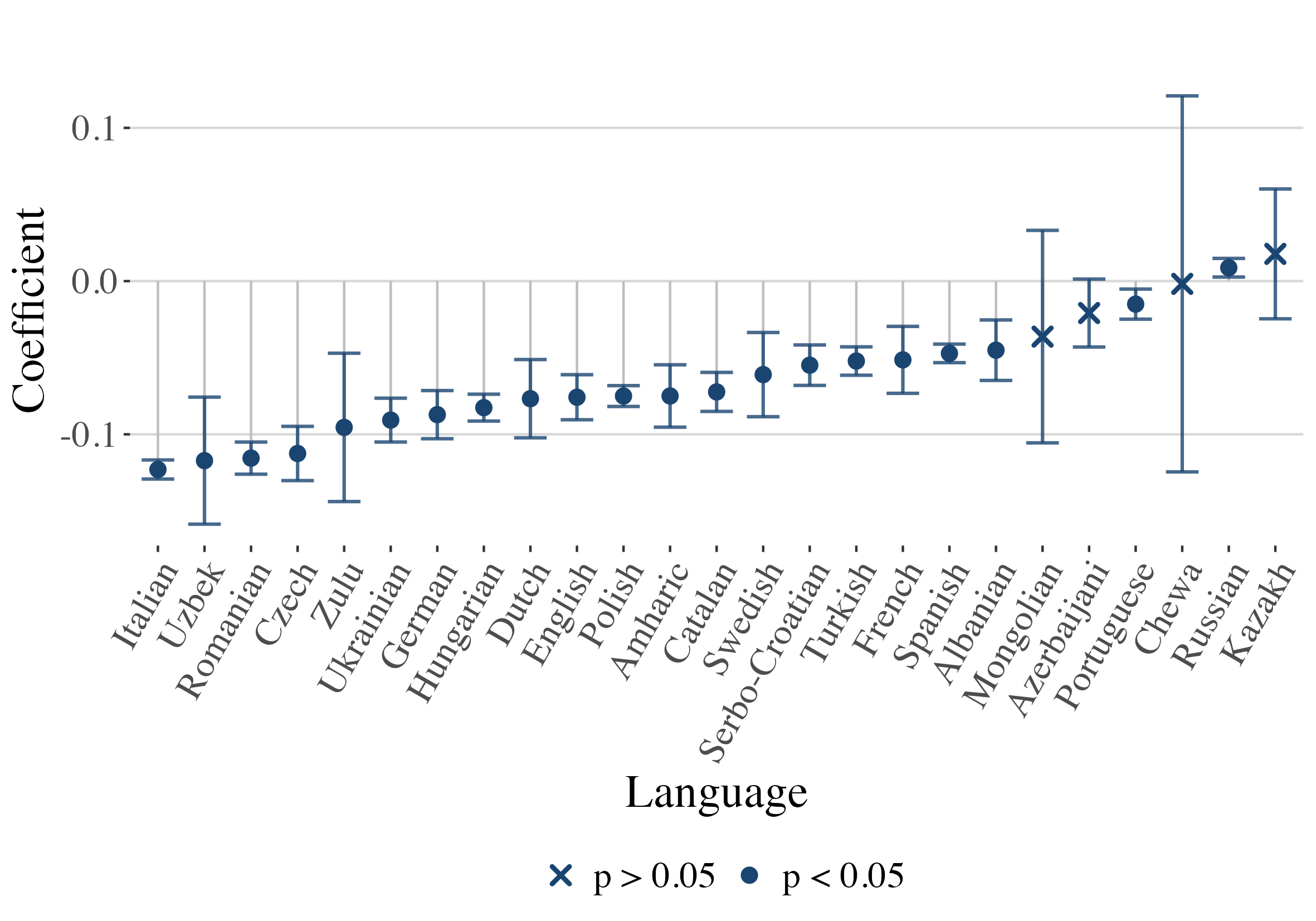}
    \caption{Phonotactic complexity and frequency regression coefficients by language, with 95\% CIs.}
    \label{fig:pc_fr_corr}
\end{figure}

Within most languages, we find a significant negative effect of frequency on phonotactic complexity after controlling for word length, as shown in \cref{fig:pc_fr_corr}. The linear model predicts an effect of $-0.06$ (95\% CI [$-0.08$, $-0.05$], $p < 0.001$, $\sigma = 0.034$). These findings are similar to those of \citet{mahowald}, who found a negative or non-significant correlation in 100\% of languages using orthographic data and a simpler measure of phonotactic complexity. Using phonetic transcriptions, we find significantly positive effects in only 1/25 languages, suggesting that this effect is fairly consistent across languages.\looseness=-1

\begin{figure*}
    \begin{subfigure}[t]{0.45\textwidth}
        \centering
        \includegraphics[width=\textwidth]{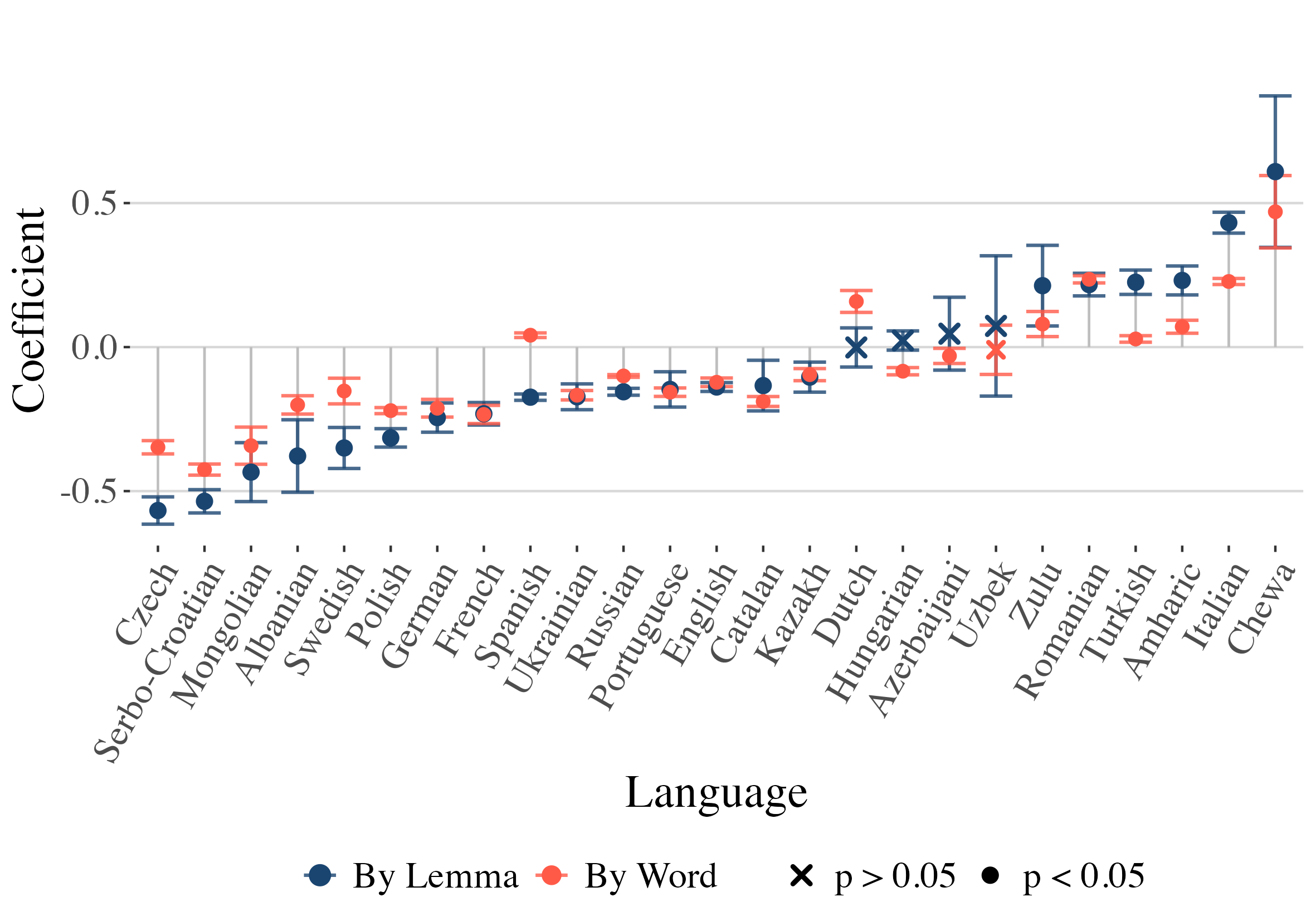}
        \caption{Regression coefficients by language, grouped by lemma and by word, with 95\% CIs.}
        \label{fig:mi_wl_corr}
        \vspace{5pt}
    \end{subfigure}
    \hfill
    \begin{subfigure}[t]{0.45\textwidth}
        \centering
        \includegraphics[width=\textwidth]{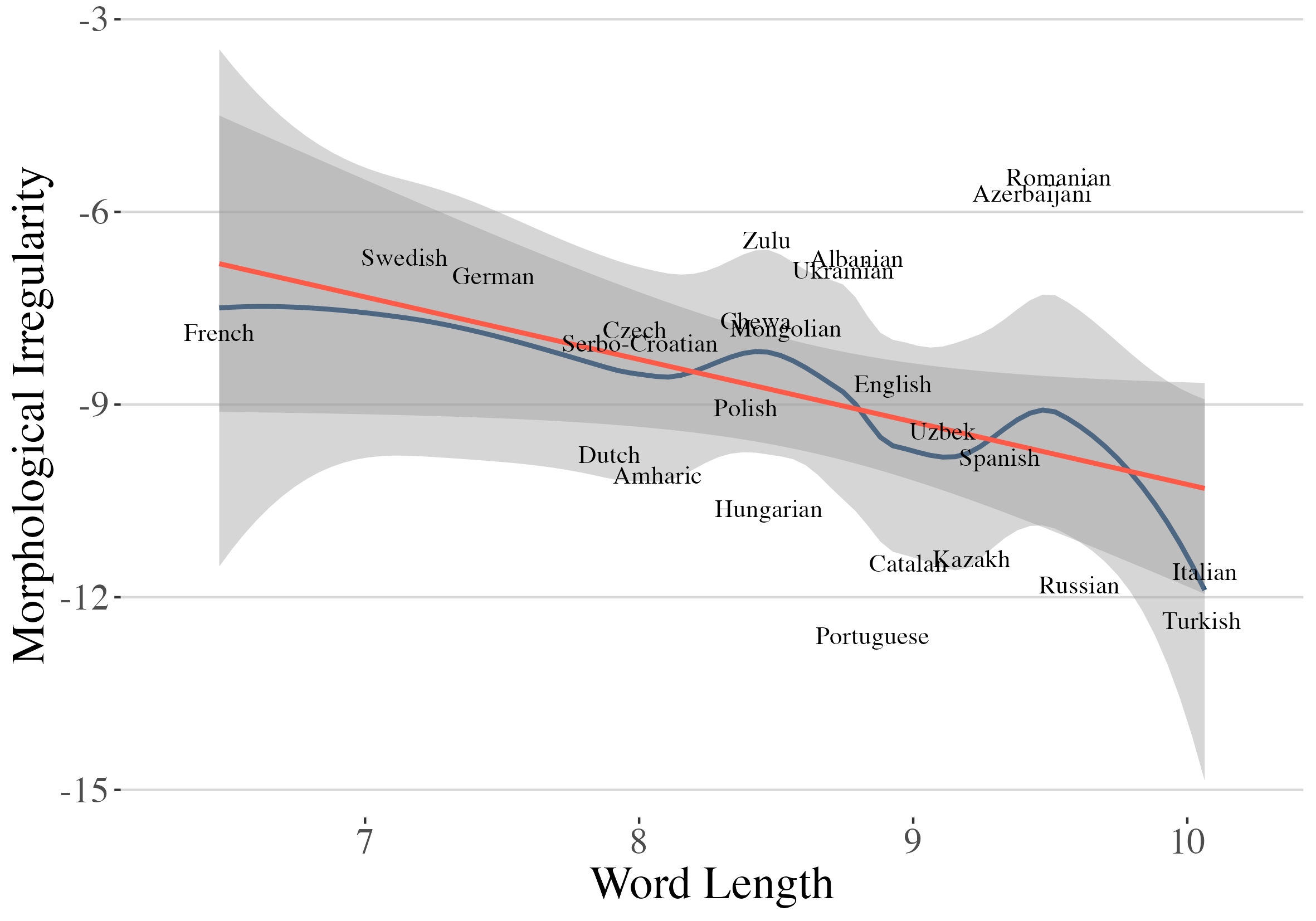}
        \caption{By-langauge means, with linear (red/light) and LOESS (blue/dark) smoothers.}
        \label{fig:mi_wl_langs}
    \end{subfigure}
    \caption{Morphological irregularity and word length.}
    \vspace{-7.5pt}
\end{figure*}

\subsection{Morphological Irregularity and Length}

Within most languages, we find a negative effect of word length on morphological irregularity after controlling for frequency, shown in \cref{fig:mi_wl_corr}. The linear model also estimates a negative effect of $-0.07$ (95\% CI [$-0.15$, 0.00], $p = 0.058$, $\sigma = 0.193$), although it is non-significant. This effect is also somewhat consistent across languages, as shown in \cref{fig:mi_wl_langs}. Across languages, we find a non-significant Spearman's correlation of $-0.38$ ($p = 0.061$), while the linear model estimates the effect of mean word length to be $-0.60$ (95\% CI [$-1.09$, $-0.12$], $p = 0.015$). When grouped by lemma, we find a non-significant effect of $-0.08$ (95\% CI [$-0.19$, 0.03], $p = 0.131$, $\sigma = 0.277$), and a non-significant effect of mean word length of $-0.59$ (95\% CI [$-1.28$, 0.11], $p = 0.098$).

\subsection{Word Length and Frequency}

\begin{figure}
    \centering
    \includegraphics[width=0.45\textwidth]{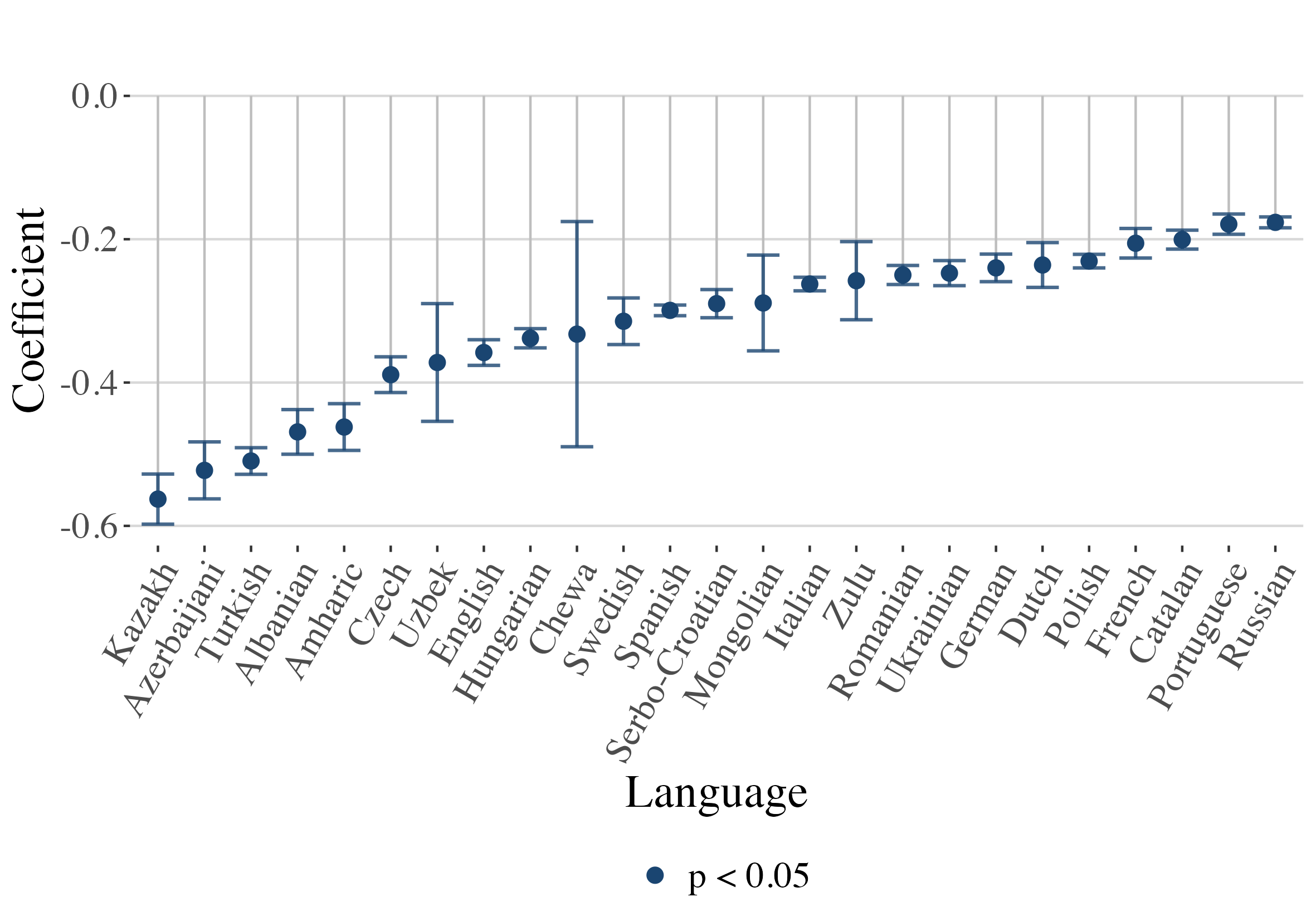}
    \caption{Word length and frequency regression coefficients by language, with 95\% CIs.}
    \label{fig:wl_fr_corr}
    \vspace{-5pt}
\end{figure}

Finally, our results for word length and frequency are exactly as expected, following the prediction of \citet{zipf}. We find consistently negative effects of frequency on word length, shown in \cref{fig:wl_fr_corr}, and an effect of $-0.32$ (95\% CI [$-0.36$, $-0.27$], $p < 0.001$, $\sigma = 0.109$) in the linear model.

\section{Discussion}

In examining the interactions between phonotactic complexity, morphological irregularity, word length, and frequency, we have found several unexpected and unintuitive results.

\paragraph{Within-language Results.}
Within most languages, we find significant effects for every pair of variables, although the direction of the effect varies. In our analysis of phonotactic complexity and frequency, we see a strong tendency towards negative effects. In our analysis of morphological irregularity and word length, we see a tendency towards a negative effect, but a positive effect in several languages. In analyses of morphological irregularity and frequency, and of phonotactic complexity and morphological irregularity, we see a strong tendency towards positive effects. However, the only analysis that is without exception in all languages examined is word length and frequency (by far the best supported by previous work, i.e., Zipf's Law), where there is a negative effect for all languages.  For phonotactic complexity and word length, we found that the direction of the effect for each language changes with the data used. These results complicate the claims of previous work examining several of these pairs across languages, which generally conclude that there is strong support for the relationship. Our results for phonotactic complexity and morphological irregularity also contradict those of \citet{hay2003phonotactics}, \citet{hay2000causes}, and \citet{burzio2002missing}. We find evidence of a positive effect of phonotactic complexity on morphological irregularity within language, rather than the negative effect that has previously been argued for. Our analysis controlled for word length and frequency, while previous work did not.

\paragraph{Across-language Results.}
Across languages, however, we find a negative effect of word length on morphological irregularity and a positive effect of word length on phonotactic complexity, although the direction of this effect changes with the data set used. We also find no evidence of a \emph{linear} effect of phonotactic complexity on morphological irregularity across languages. However, as can be seen in \cref{fig:mi_pc_langs}, there may be a nonlinear effect across languages. In exploratory data analysis, we also identified several possible nonlinear effects within languages.
We leave fully describing any such relationships to future work.

\paragraph{Incomplete Picture.}
The results presented here suggest that, although these four variables do influence each other, we do not have enough information to make claims about one compensating for another, either within a single language or universally across languages. The effects found are potentially consistent with several causal models. It is the causal effects captured by these models that we are interested in---how much does an intervention in one variable \textit{directly} affect another, if at all? If a decrease in morphological irregularity causes a decrease in word length, which then causes an increase in phonotactic complexity, there is no compensatory relationship between morphological irregularity and phonotactic complexity, no matter how correlated they appear to be.

\paragraph{Causal Modeling.}
Our results assume that the underlying causal structure implied by previous work is correct---that morphological irregularity and phonotactic complexity have a common cause of frequency and word length, that the effect of frequency on phonotactic complexity is mediated by word length, and that morphological irregularity and word length have a common cause of frequency. However, it is possible that a different causal model underlies this data. We leave proposing such causal models and testing their implications for future work. While the models discussed in this work provide a starting point for understanding the structure of the lexicon, evidence supporting the underlying causal structure responsible for generating the data is necessary to evaluate any compensatory relationships in a set of highly correlated variables \citep{pearl2016causal}.\looseness=-1

\section*{Acknowledgements}

We thank the Montreal Computational \& Quantitative Linguistics Lab for helpful feedback. The first author was supported by funding from the Fonds de recherche du Québec - Société et culture (FRQSC). The third author acknowledges the support of the Natural Sciences and Engineering Research Council of Canada (RGPIN-2023-04873). The senior author also gratefully acknowledges the support of the Natural Sciences and Engineering Research Council of Canada  and the Canada CIFAR AI Chairs Program. This research was enabled in part by support provided by Calcul Québec  and the Digital Research Alliance of Canada.\looseness=-1

\bibliography{custom, anthology}

\vfill\clearpage

\appendix
\onecolumn
\section{Languages in Analysis}
\label{sec:appendix_a}

\begin{table*}[h!]
    \centering
    \begin{tabular}{lllrrlrr}
        \hline
        \textbf{Language}   & \textbf{Family} & \textbf{Type} & \multicolumn{1}{l}{\textbf{Forms}} & \multicolumn{1}{l}{\textbf{Lemmas}} & \textbf{Phon.} & \multicolumn{1}{l}{\textbf{Acc.}} \\ \hline
        Albanian            & Indo-Eur.       & Fus.          & 33483                              & 589                                 & NL             & 81.16                             \\
        Amharic             & Afro-Asiatic    & Fus.          & 46224                              & 2461                                & WP             & 92.82                             \\
        Azerbaijani         & Turkic          & Agg.          & 8004                               & 340                                 & NL             & 76.15                             \\
        Catalan             & Indo-Eur.       & Fus.          & 81576                              & 1547                                & NL             & 95.88                             \\
        Chewa               & Atl. Congo      & Agg.          & 4370                               & 227                                 & WP             & 85.91                             \\
        Czech               & Indo-Eur.       & Fus.          & 50284287                           & 824074                              & NL             & 89.89                             \\
        Dutch               & Indo-Eur.       & Fus.          & 55467                              & 4993                                & NL             & 92.33                             \\
        English             & Indo-Eur.       & Fus.          & 115523                             & 22765                               & NL             & 80.89                             \\
        French              & Indo-Eur.       & Fus.          & 367732                             & 7535                                & NL             & 84.96                             \\
        German              & Indo-Eur.       & Fus.          & 179339                             & 15060                               & NL             & 86.95                             \\
        Hungarian           & Uralic          & Agg.          & 490394                             & 13989                               & NL             & 93.27                             \\
        Italian             & Indo-Eur.       & Fus.          & 509574                             & 10009                               & NL             & 98.85                             \\
        Kazakh              & Turkic          & Agg.          & 40283                              & 1755                                & NL             & 98.00                             \\
        Khalka Mongolian    & Mongolic        & Agg.          & 30143                              & 2140                                & NL             & 82.35                             \\
        Polish              & Indo-Eur.       & Fus.          & 13882543                           & 274550                              & NL             & 90.93                             \\
        Portuguese          & Indo-Eur.       & Fus.          & 303996                             & 4001                                & NL             & 97.59                             \\
        Romanian            & Indo-Eur.       & Fus.          & 80266                              & 4405                                & NL             & 78.09                             \\
        Russian             & Indo-Eur.       & Fus.          & 473481                             & 28068                               & NL             & 95.53                             \\
        Serbo-Croatian      & Indo-Eur.       & Fus.          & 840799                             & 24419                               & NL             & 92.30                             \\
        Spanish             & Indo-Eur.       & Fus.          & 382955                             & 5460                                & NL             & 83.35                             \\
        Swedish             & Indo-Eur.       & Fus.          & 78411                              & 10553                               & NL             & 83.23                             \\
        Turkish             & Turkic          & Agg.          & 275460                             & 3579                                & NL             & 96.07                             \\
        Ukranian            & Indo-Eur.       & Fus.          & 20904                              & 1493                                & NL             & 85.11                             \\
        Uzbek               & Turkic          & Agg.          & 810                                & 68                                  & WP             & 87.86                             \\
        Zulu                & Atl. Congo      & Agg.          & 49562                              & 621                                 & WP             & 75.45                             \\
        \hline
        \textit{Bengali}    & Indo-Eur.       & Fus.          & 4443                               & 136                                 & NL             & 37.75                             \\
        \textit{Cebuano}    & Austronesian    & Agg.          & 618                                & 97                                  & WP             & 54.17                             \\
        \textit{Hindi}      & Indo-Eur.       & Fus.          & 54438                              & 258                                 & NL             & 68.79                             \\
        \textit{Indonesian} & Austronesian    & Agg.          & 27714                              & 3877                                & WP             & 49.26                             \\
        \textit{Kabardian}  & Abkhaz-Adyge    & Agg.          & 3092                               & 250                                 & WP             & 63.36                             \\
        \textit{Kashubian}  & Indo-Eur.       & Fus.          & 509                                & 37                                  & WP             & 1.43                              \\
        \textit{Kyrgyz}     & Turkic          & Agg.          & 5544                               & 98                                  & WP             & 29.81                             \\
        \textit{Maltese}    & Afro-Asiatic    & Fus.          & 3584                               & 112                                 & WP             & 29.82                             \\
        \textit{Swahili}    & Atl. Congo      & Fus.          & 14130                              & 185                                 & WP             & 35.17                             \\
        \textit{Tagalog}    & Austronesian    & Agg.          & 2912                               & 344                                 & WP             & 46.22                             \\
        \textit{Tajik}      & Indo-Eur.       & Fus.          & 77                                 & 75                                  & WP             & 27.01                             \\
        \textit{Telugu}     & Dravidian       & Agg.          & 1548                               & 127                                 & NL             & 41.67                             \\
        \textit{Turkmen}    & Turkic          & Agg.          & 810                                & 68                                  & WP             & 30.42                             \\
        \textit{Urdu}       & Indo-Eur.       & Fus.          & 12572                              & 182                                 & WP             & 50.89                             \\
        \textit{Uyghur}     & Turkic          & Agg.          & 8178                               & 90                                  & WP             & 15.12                             \\         \hline
    \end{tabular}
    \caption{All languages in analysis. \textit{Italicized} languages are excluded. Abbreviations: Fus. - Fusional; Agg. - Agglutinative; Phon. - Phonotactic data source; NL - NorthEuraLex; WP - WikiPron; Acc. - Morphology Model Accuracy}
\end{table*}
\vfill\clearpage

\section{Plots of Individual Languages}
\label{sec:appendix_b}

\begin{figure}[htb!]
    \begin{subfigure}{0.45\textwidth}
        \centering
        \includegraphics[width=\textwidth]{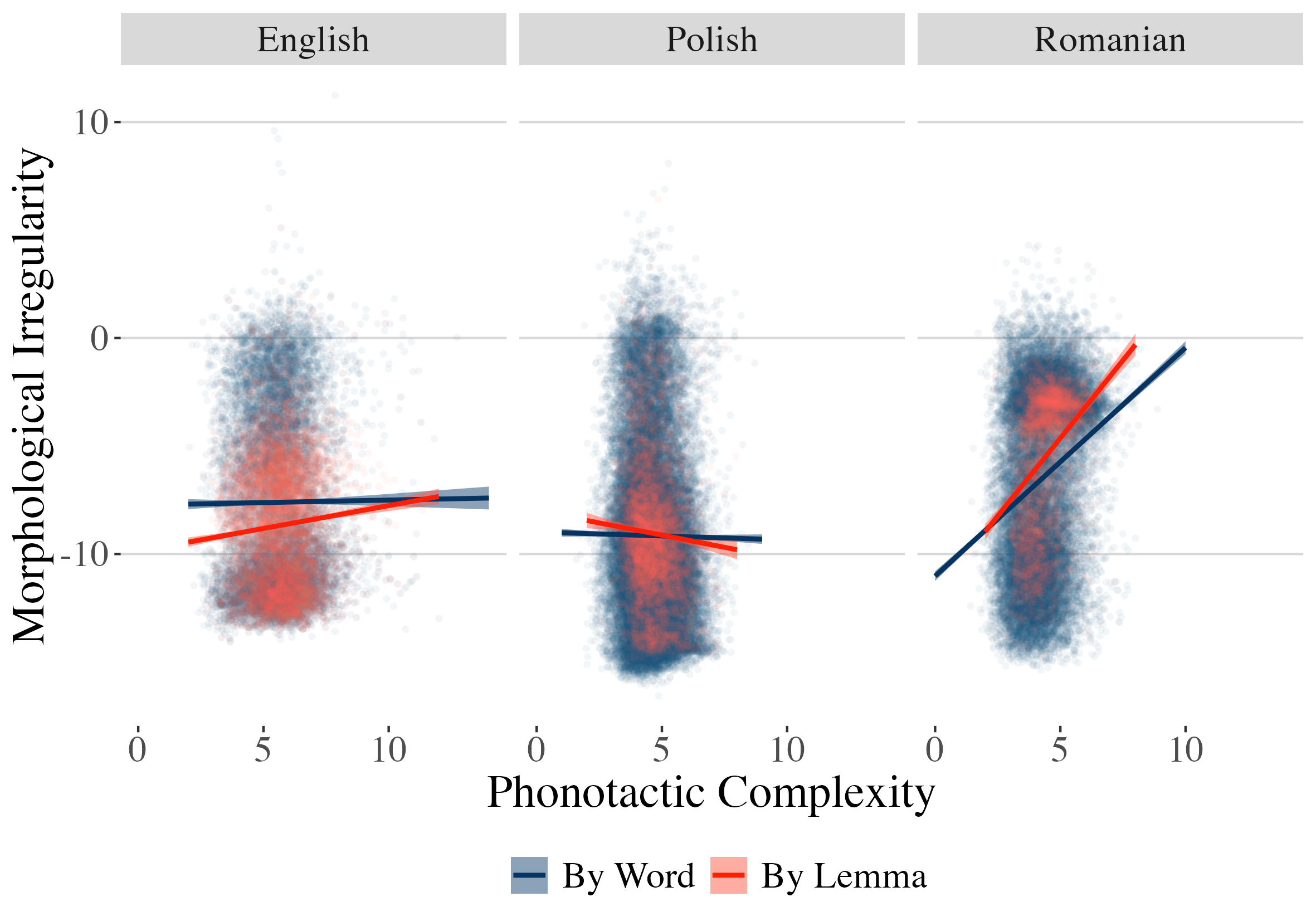}
        \caption{Morphological irregularity and phonotactic complexity data for English, Polish, and Romanian, with linear model predictions from \cref{tab:regression} and 95\% CIs.}
        \vspace{40pt}
    \end{subfigure}
    \hfill
    \begin{subfigure}{0.45\textwidth}
        \centering
        \includegraphics[width=\textwidth]{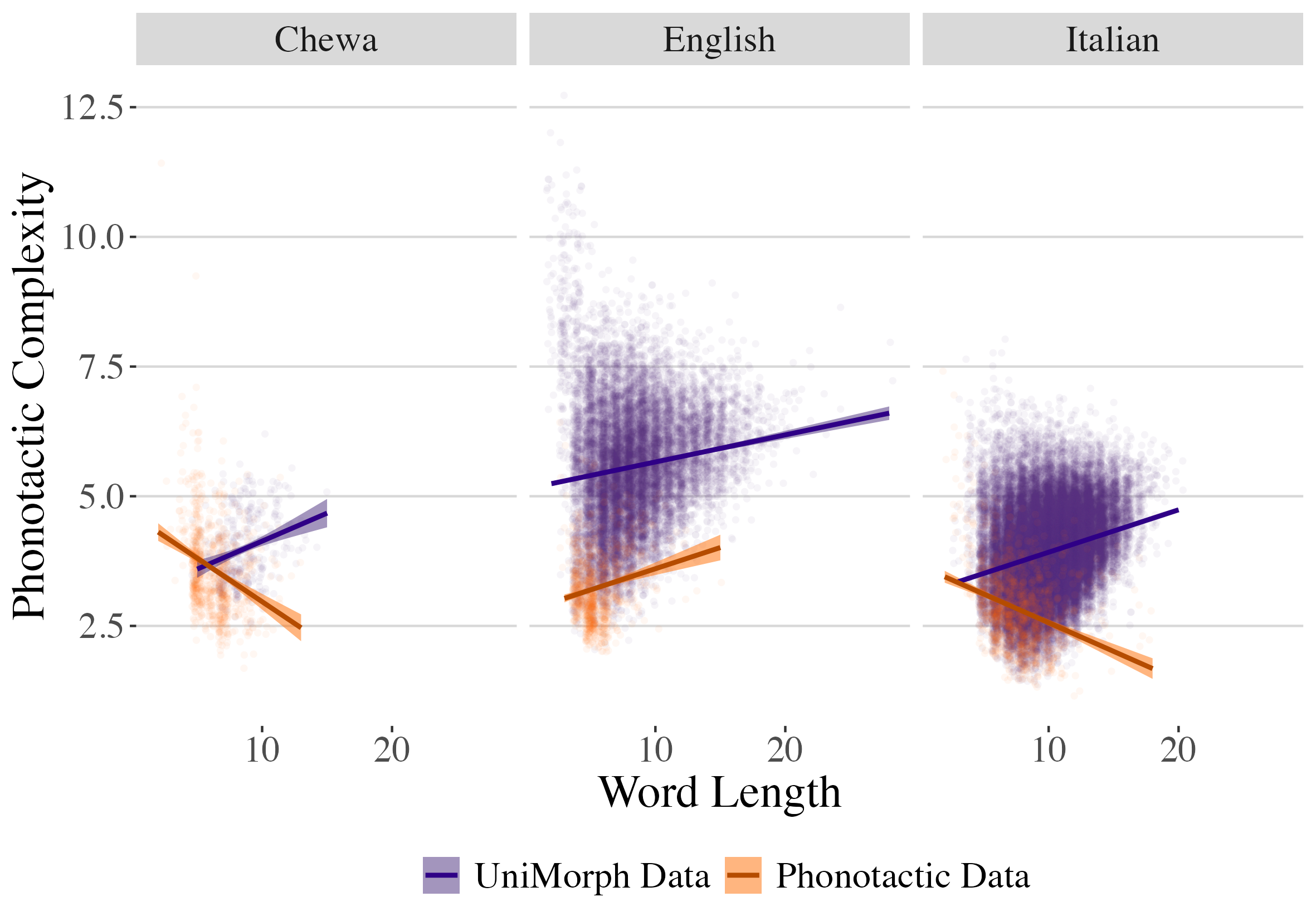}
        \caption{Phonotactic complexity and word length data for Chewa, English, and Italian, with linear model predictions from \cref{tab:regression} and 95\% CIs.}
        \vspace{40pt}
    \end{subfigure}

    \begin{subfigure}{0.45\textwidth}
        \centering
        \includegraphics[width=\textwidth]{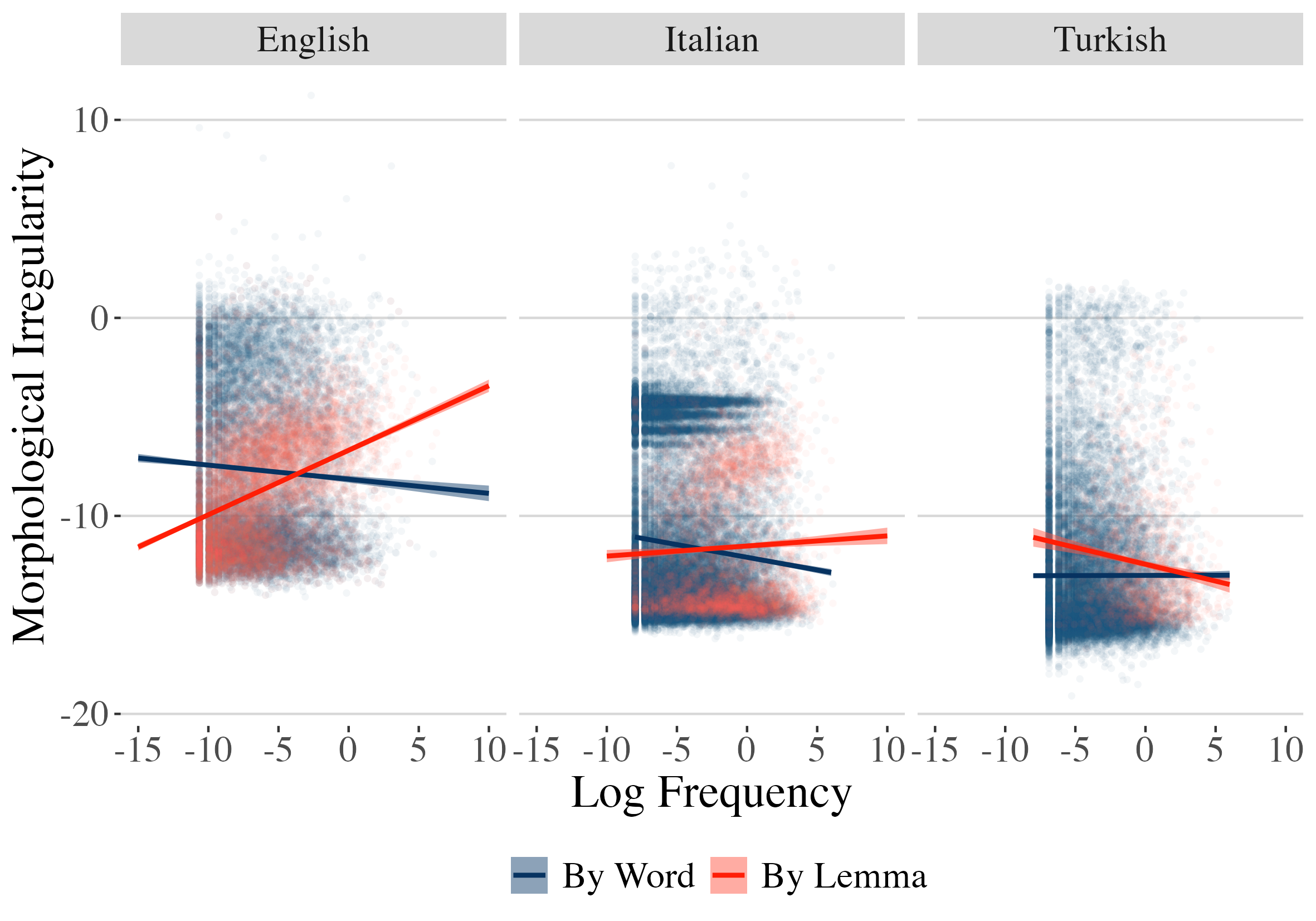}
        \caption{Morphological irregularity and frequency data for English, Italian, and Turkish, with linear model predictions from \cref{tab:regression} and 95\% CIs.}
        \vspace{40pt}
    \end{subfigure}
    \hfill
    \begin{subfigure}{0.45\textwidth}
        \centering
        \includegraphics[width=\textwidth]{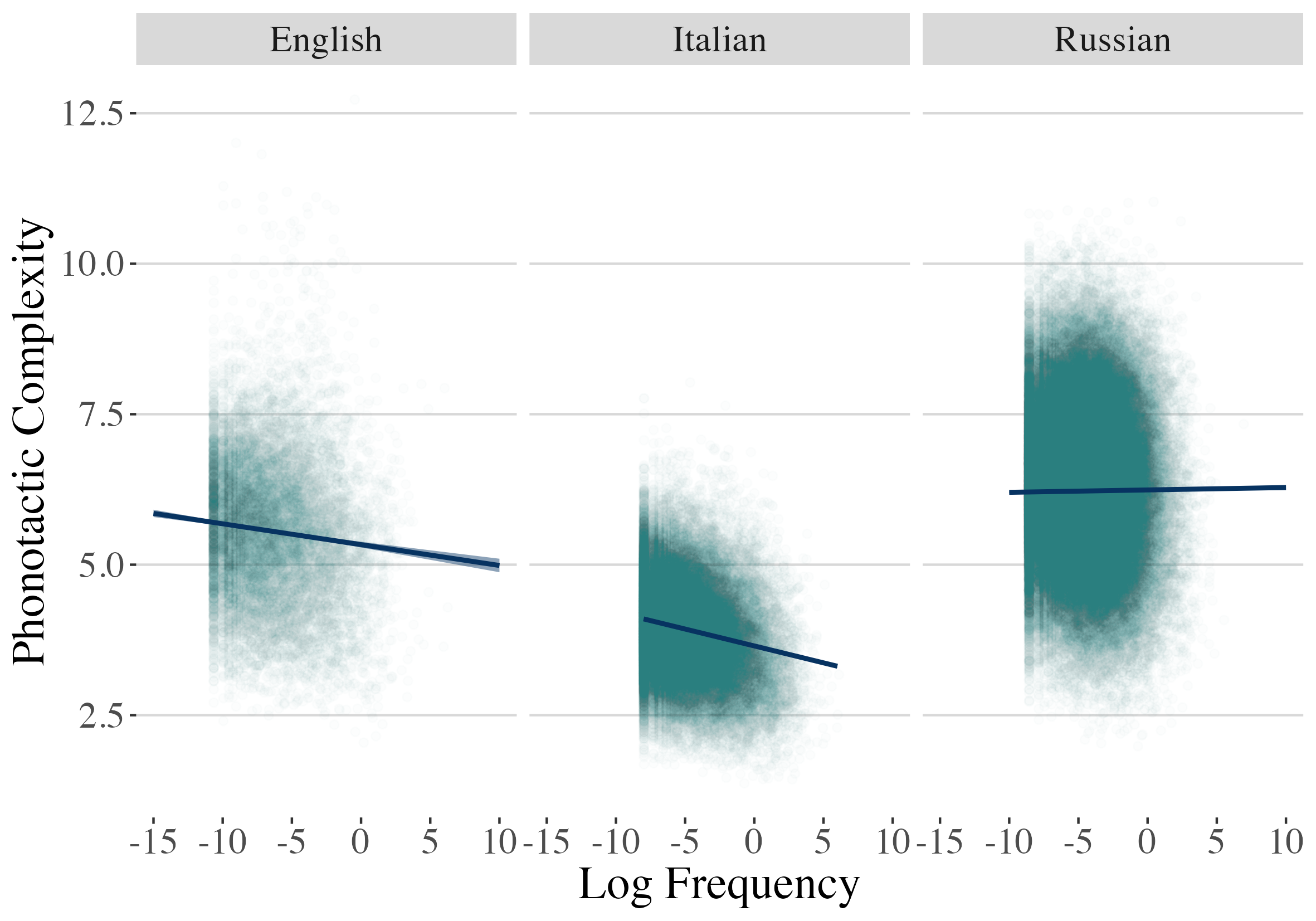}
        \caption{Phonotactic complexity and frequency data for English, Italian, and Russian, with linear model predictions from \cref{tab:regression} and 95\% CIs.}
        \vspace{40pt}
    \end{subfigure}

    \begin{subfigure}{0.45\textwidth}
        \centering
        \includegraphics[width=\textwidth]{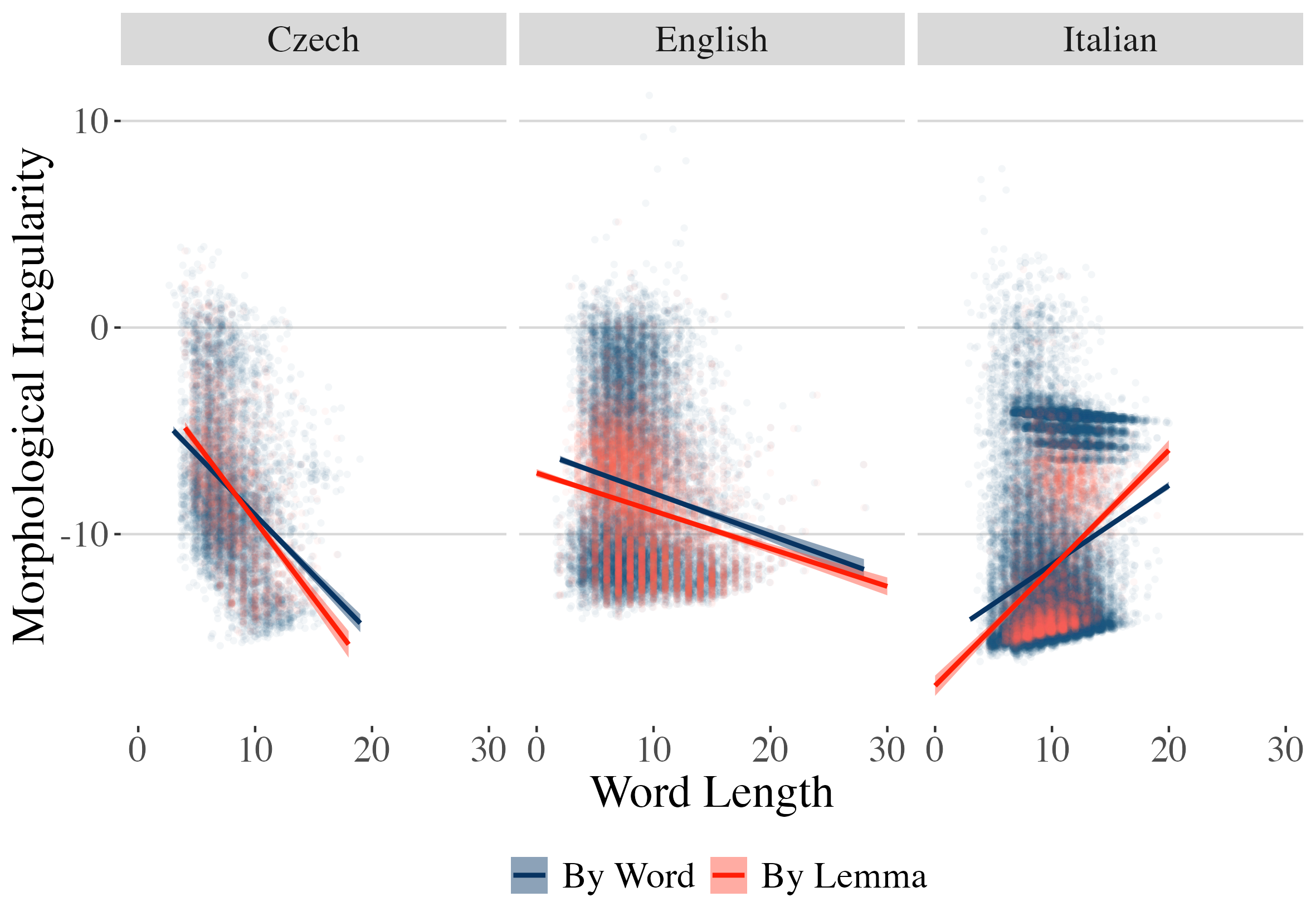}
        \caption{Morphological irregularity and word length data for Czech, English, and Italian, with linear model predictions from \cref{tab:regression} and 95\% CIs.}
    \end{subfigure}
    \hfill
    \begin{subfigure}{0.45\textwidth}
        \centering
        \includegraphics[width=\textwidth]{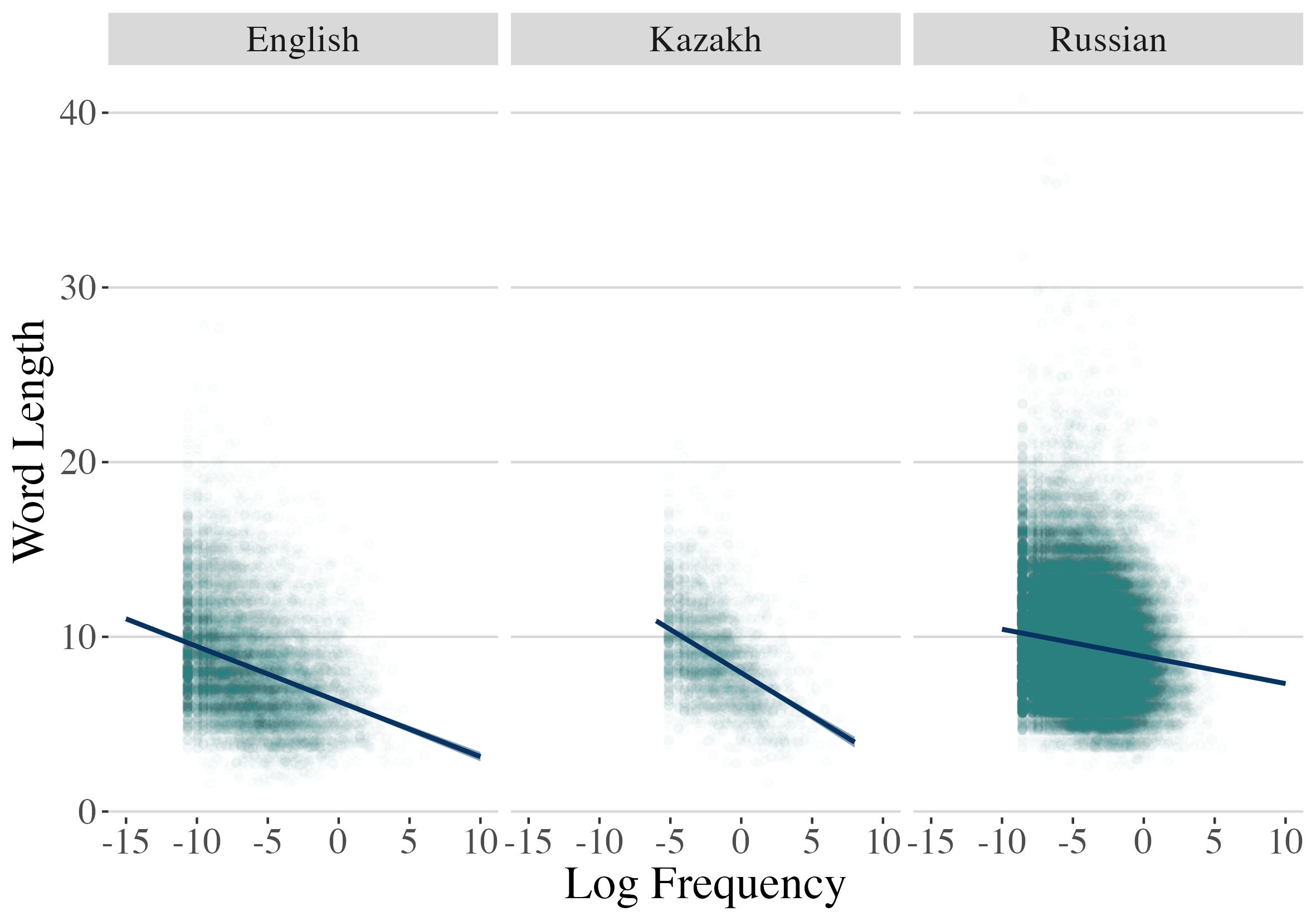}
        \caption{Word Length and frequency data for English, Kazakh, and Russian, with linear model predictions from \cref{tab:regression} and 95\% CIs.}
    \end{subfigure}
\end{figure}

\end{document}